\documentclass[journal, 10pt,twocolumn]{IEEEtran}

\usepackage{amssymb}
\usepackage[all,arc]{xy}
\usepackage{mathrsfs}
\usepackage{algorithm2e}
\usepackage{amsbsy}
\usepackage{amsfonts}
\usepackage{amsmath}
\usepackage{amssymb}
\usepackage{amsthm}
\usepackage{array}
\usepackage{authblk}

\usepackage{bbm} 
\usepackage[font=footnotesize,labelfont=bf]{caption}
\usepackage{color}
\usepackage[mathscr]{eucal}
\usepackage{mathrsfs}

\usepackage{float}
\usepackage{hyperref} 
\usepackage{stmaryrd}
\usepackage[font=scriptsize,labelfont=scriptsize]{subcaption}
\usepackage{graphicx}
\usepackage{url}
\usepackage{verbatim}
\usepackage{adjustbox}
\usepackage[usenames,dvipsnames,svgnames,table]{xcolor}
\usepackage{multirow,array}

\usepackage{tikz}
\newcommand\copyrighttext{%
  \footnotesize \textcopyright \ 2018 IEEE.  Final accepted version should be cited as: Czaja, W., J.M. Murphy, and D. Weinberg. ``Superresolution of Noisy Remotely Sensed Images Through Directional Representations." IEEE Geoscience and Remote Sensing Letters 99 (2018): 1-5.  DOI: 10.1109/LGRS.2018.2865131.}
\newcommand\copyrightnotice{%
\begin{tikzpicture}[remember picture,overlay]
\node[anchor=south,yshift=10pt] at (current page.south) {\fbox{\parbox{\dimexpr\textwidth-\fboxsep-\fboxrule\relax}{\copyrighttext}}};
\end{tikzpicture}%
}

\makeatletter
\def\ps@pprintTitle{%
 \let\@oddhead\@empty
 \let\@evenhead\@empty
 \def\@oddfoot{}%
 \let\@evenfoot\@oddfoot}
\makeatother

\title{Superresolution of Noisy Remotely Sensed Images Through Directional Representations}

\author[a]{Wojciech Czaja}
\author[b]{James M. Murphy}
\author[c]{Daniel Weinberg}

\affil[a]{University of Maryland, Department of Mathematics, College Park, MD}

\affil[b]{Johns Hopkins University, Department of Mathematics, Baltimore, MD}

\affil[c]{US Census Bureau, Center for Statistical Research and Methodology, Washington, DC}

\begin{document}

\maketitle

\begin{abstract}We develop an algorithm for single-image superresolution of remotely sensed data, based on the discrete shearlet transform.  The shearlet transform extracts directional features of signals, and is known to provide near-optimally sparse representations for a broad class of images.  This often leads to superior performance in edge detection and image representation when compared to isotropic frames.  We justify the use of shearlets mathematically, before presenting a denoising single-image superresolution algorithm that combines the shearlet transform with sparse mixing estimators (SME).  Our algorithm is compared with a variety of single-image superresolution methods, including wavelet SME superresolution.  Our numerical results demonstrate competitive performance in terms of PSNR and SSIM.  
\end{abstract}

\copyrightnotice

\section{Introduction}

Superresolution is the problem of increasing the resolution of an image without introducing artifacts.  It is a significant problem in image processing generally \cite{Park2003} and remote sensing specifically \cite{Gu2008, VillaChanussot2010, Zhang2012, Liu2014, Li2016}.  We aim to recover an image signal $f:[0,1]^{2}\rightarrow\mathbb{R}$ given measurements \begin{align}\label{problem}y=\mathcal{L}(f)+\mathcal{N},\end{align}where $\mathcal{L}$ is a degradation operator, for example a low-pass filter, and $\mathcal{N}$ is a noise term.  In our case, $\mathcal{L}(f)$ is a downsampled version of $f$, $\mathcal{N}$ is a noise image where each pixel is generated as a realization of a Gaussian random variable,  and $y$ is a noisy and downsampled version of $f$.  In this context, the goal of superresolution is to recover the original image $f$ by increasing the resolution and denoising $y$.  Qualitative analysis of superresolution is based on visual appearance of the superresolved image, which should be sharp and without artifacts.  Quantitative evaluation measures include peak-signal-to-noise-ratio (PSNR) and the structural similarity index metric (SSIM) \cite{Wang2004}, which is designed to be consistent with human visual perception.

The present article develops an algorithm motivated by the rich theory of anisotropic harmonic analysis.  We consider a generalization of  the state-of-the-art method of superresolution with wavelet sparse mixing estimators (SME).  This method was introduced by S. Mallat and G. Yu \cite{Mallat2010}, and performs strongly in terms of visual quality and quantitative metrics.  Our generalization incorporates anisotropic shearlets into this regime, in order to capitalize on the theoretical near-optimality of shearlets for sparsely representing certain classes of images \cite{Guo2007, compact_shearlets}.

\section{Background on Superresolution}\label{sec:sr}

\subsection{Problem Formulation and Basic Methods}

Superresolution aims to increase the resolution of the data $y$ by recovering $f$ as in (\ref{problem}).  The signal produced by the superresolution algorithm may then be compared to the original image before degrading and adding noise.  Theoretically, this involves solving an inverse problem (\ref{problem}) that is usually ill-posed.  For numerical algorithms, interpolation is usually performed instead of inversion.  Letting $I$ be a discrete image represented as an $M\times N$ matrix, a superresolution algorithm outputs an $\tilde{M}\times\tilde{N}$ matrix $\tilde{I}$ , with $M<\tilde{M}, \ N<\tilde{N}$.  We consider a typical case where $\tilde{M}=2M$ and $\tilde{N}=2N$, corresponding to doubling the resolution of the original image.  Images with multiple channels, such as hyperspectral images, can be superresolved by superresolving each channel separately.

Superresolution can be implemented by using information in addition to $I$, such as low resolution images at sub-pixel shifts of the scene \cite{SR_overview}, or images of the scene with different modalities.  The latter method is related to the specific problem of pan-sharpening \cite{Czaja2014}.  Alternatively, superresolution can be performed using only $I$; this is called \emph{single-image superresolution}.  In this article, we develop a single-image superresolution method requiring as input only the image $I$, along with relevant algorithm parameters.

\subsection{Superresolution in Remote Sensing}

A variety of techniques for superresolution of remotely sensed data have been developed, including spatial-spectral methods \cite{Gu2008}, neural networks \cite{Tatem2001}, and sparse dictionary methods \cite{Peleg2014}. The superresolution of remotely sensed hyperspectral data has been used to improve classification performance, where low spatial resolution leads to mixed pixels for which classification is very difficult, \cite{VillaChanussot2010}.  Different superresolution methods have successfully improved the accuracy of waterline mappings \cite{Foody2005} and land cover change \cite{Li2016}.  More general problems in target identification using remotely sensed data have also been addressed via superresolution algorithms \cite{Tatem2001}.

In addition, the superresolution of general remote sensing data is significant for image registration.  The registration of multimodal remote sensing images is a significant problem, and a particular challenge is that images to be registered are sometimes not of the same resolution \cite{Murphy2016}.  A common solution is to downsample the higher resolution data until it is the same resolution as the lower resolution data \cite{Zavorin2005, Murphy2016}.  This, however, destroys valuable information via downsampling.  It is more desirable superresolve the lower resolution data, which preserves the detailed information in the high resolution image.  Registering multimodal data is particularly challenging, so we consider our algorithm on a variety of image modalities.  The proposed algorithm performs purely spatial single-image superresolution, and does not superresolve along the spectral dimension of multispectral and hyperspectral images.  Instead, each band is superresolved individually.  

\subsection{Superresolution with Sparse Mixing Estimators}

One method that has achieved state-of-the-art results is superresolution with sparse mixing estimators (SME) \cite{Mallat2010}.  This algorithm, introduced by S. Mallat and G. Yu, takes advantage of block sparsity \cite{Eldar_block} by decomposing the image to be superresolved into a redundant representation, then directionally interpolating based on a sparse frame representation.

Given a discrete signal $y$ as in (\ref{problem}), we wish to represent $y$ according to a sparse mixture model.  To do this, we consider a family of blocks $\mathcal{B}$ and a representation system $\Psi$.  The family of blocks may be understood as subsets of the space of all coefficients.  Visually, these are small parallelograms in the coefficient domain.  Our representation system is an example of a frame, which generalizes the notion of orthonormal basis to allow for redundant representations \cite{Christensen2003}.

The coefficients of $y$ with respect to $\Psi$ are denoted $c$.  The operator taking $y$ as input and outputting the coefficients of $y$ in the representation system $\Psi$ is called the analysis operator, and is denoted $\mathcal{F}_{\Psi}$, so that $\mathcal{F}_{\Psi}y=c$.  The synthesis operator $\mathcal{F}_{\Psi}^{*}$ reconstructs the original signal $y$ from its coefficients $c$, so that $\mathcal{F}_{\Psi}^{*}c= y$ in the case that $\Psi$ is a tight frame.  We let $y_{B}=\mathcal{F}_{\Psi}^{*}(c\mathbbm{1}_{B})$ be the synthesis operator applied only to the block of coefficients $B$, where $\mathbbm{1}_{B}$ is the indicator function for the set $B$.  Then we seek to write $y$ in the form: \begin{align}\label{SME_decomp}y=\sum_{B\in\mathcal{B}}\tilde{a}(B)y_{B}+y_{r},\end{align}for a choice of mixing coefficients $\{\tilde{a}(B)\}_{B\in\mathcal{B}}$ and a residual term $y_{r}$.  The mixing coefficients determine how strongly the coefficients from a given block will contribute to the reconstructed signal.  Because $\Psi$ is redundant, many choices of $\{\tilde{a}(B)\}_{B\in\mathcal{B}}$ will satisfy (\ref{SME_decomp}).  The coefficients should be chosen for both fidelity and to promote block sparsity, which will be used to efficiently directionally interpolate the image, hence: 

\begin{align}\label{CoeffOptimization}\tilde{a}=&\arg\min_{a}\frac{1}{2}\left\|c\left(1-\sum_{B\in\mathcal{B}}a(B)\mathbbm{1}_{B}\right)\right\|_{2}^{2}\\+&\lambda\sum_{B\in\mathcal{B}}|a(B)|\|\bar{R}_{B}c\|_{B}^{2},\nonumber \end{align}where $\|\cdot\|_{B}^{2}$ is the squared $\ell^{2}$ norm on the block $B$ and $\bar{R}_{B}c=c|_{B}-\bar{c}|_{B}$, where, $\bar{c}|_{B}(k,j)=$ average of the $k^{th}$ representation coefficients in block $B$ located on the line passing through $j$, at angle $\theta$, where $k$ runs through all representation coefficients and $\theta$ is the orientation of $B$.  The first term of (\ref{CoeffOptimization}) is a fidelity term while the second term is a regularizer that enforces block sparsity.  Intuitively, if $\|\bar{R}_{B}c\|_{B}^{2}$ is small, then there is a strong degree of directional regularity present in the block, which can be used for directional interpolation.  Such blocks are favored by (\ref{CoeffOptimization}).  The fidelity and sparsity components are balanced by a tunable regularization parameter $\lambda$.  

The signal of interest, $f$, is then estimated via the decomposition (\ref{SME_decomp}).  Let $\mathcal{B}_{\theta}$ be the set of blocks oriented in the direction $\theta$, so that $\mathcal{B}=\bigcup_{\theta}\mathcal{B}_{\theta}.$  Let $U_{\theta}^{+}$ be a directional interpolator in the direction $\theta$, and $U^{+}$ an isotropic interpolator.  We estimate $f$ as $\tilde{f}=U^{+}y+\sum_{\theta}(U_{\theta}^{+}-U^{+})\tilde{\Psi}\left(\sum_{B\in\mathcal{B}_{\theta}}\tilde{a}(B)c\mathbbm{1}_{B}\right),$ where $\tilde{\Psi}$ is the dual of $\Psi$; see \cite{Christensen2003}, Chapter 4.  This simultaneously denoises and superresolves $y$.

This paper develops a single-image superresolution algorithm that computes dominant directions efficiently and accurately, using the harmonic analytic construction of \emph{shearlets} \cite{easley+labate+lim,  Guo2007}.  This method is quite general, can be applied to images of any size, and has few tunable parameters.   Moreover, shearlets are known to provide near-optimally sparse representations for models of large classes of natural images.  The algorithm presented in this article builds upon an elementary prototype \cite{MurphyPhD2015} by more efficiently utilizing shearlet sparsity.

\section{Background on Harmonic Analysis}\label{sec:math}

Shearlets generalize wavelets by incorporating a notion of directionality.  We thus begin our mathematical discussion of shearlets with background on wavelets \cite{Daubechies1992, Mallat1999}.  

In a broad sense, wavelet algorithms decompose an image with respect to \emph{scale} and \emph{translation}.  Mathematically, given a signal $f\in L^{2}([0,1]^{2})$, understood as a continuous image signal, and an appropriately chosen wavelet function $\psi$, $f$ may be written as $f = \sum_{m\in\mathbb{Z}}\sum_{n\in\mathbb{Z}^{2}} \langle f, \psi_{m,n}\rangle \psi_{m,n},$ where:
$\psi_{m,n}(x):=|\det A|^{m/2} \psi(A^{m}x-n)$ and $A$ is an invertible $2\times 2$ matrix.  A typical choice for $A$ is the dyadic isotropic matrix $A=\left( \begin{array}{cc} 2 & 0  \\ 0 & 2\\ \end{array} \right).$ The wavelet coefficients $\{\langle f, \psi_{m,n}\rangle\}_{m\in\mathbb{Z} ,n\in\mathbb{Z}^{2}}$ describe the multiscale behavior of $f$.  This infinite scheme is truncated to work with real, finite image signals.  Wavelets have been applied to image compression \cite{Mallat1999}, image fusion \cite{Pajares2004},  and image registration \cite{Zavorin2005}.  

Shearlets generalize the multiresolution character of wavelets by decomposing with respect not just to \emph{scale} but also \emph{direction}.  Mathematically, given a signal $f\in L^{2}([0,1]^{2})$ and an appropriate shearlet function $\psi$, we may decompose $f$ as $f = \sum_{i\in\mathbb{Z}}\sum_{j\in \mathbb{Z}}\sum_{k\in\mathbb{Z}^{2}}\langle f, \psi_{i,j,k}\rangle \psi_{i,j,k},$where: $\psi_{i,j,k}(x)= \ 2^{\frac{3i}{4}}\psi(B^{j}A^{i}x-k),$ $A=\left( \begin{array}{cc} 2 & 0  \\ 0 & 2^{\frac{1}{2}}\nonumber \\ \end{array} \right), \ B=\left( \begin{array}{cc} 1 & 1  \\ 0 & 1 \\ \end{array} \right).$  Note that $A$ is anisotropic, hence it will allow our new analyzing functions to be more pronounced in a particular direction.  The new matrix $B$, a shearing matrix, selects the direction.  The shearlet coefficients $\{\langle f, \psi_{i,j,k}\rangle\}_{i,j\in\mathbb{Z}, k\in\mathbb{Z}^{2}}$ describe the behavior of $f$ at different scales (determined by $i$) and directions (determined by $i, j$).  Higher dimensional notions of shearlets exist as well \cite{CzajaKing2012, CzajaKing2014, Negi2012}.  

Theoretically, shearlets provide near-optimally sparse representations for models of large classes of natural images \cite{Guo2007}, in particular, for images that are smooth except for a finite collection of smooth discontinuities.  Practically, such images are those that are smooth except for smooth edges, which allows for the efficient use of anisotropic shearlets for a variety of image processing tasks, including image denoising \cite{Easley2009}, image registration \cite{ Murphy2016, Murphy2015, Murphy2016_Agile}, and image fusion \cite{Miao2011}.  Remotely sensed images often have significant directional features, which suggests the efficacy of shearlets for analysis of remotely sensed images.  Preliminary results of shearlet superresolution have been reported \cite{CzajaMurphy2015, Bosch2015}.  

\section{Description of Algorithm}\label{sec:alg}

Given an image $I$ to be superresolved, our algorithm first decomposes the image in a redundant shearlet frame.  Then, the shearlet coefficients are used to determine regions of the image to directionally superresolve; this step involves sparse mixing estimators.  This locally superresolves different parts of the image according to local directionality.  We call our method \emph{shearlet sparse mixing estimators (SSME)}.  In detail, the algorithm proceeds as follows.

\begin{enumerate}

\item Decompose $I$ in a redundant shearlet frame to acquire coefficients $c=\mathcal{SH}(I)$, where $\mathcal{SH}$ is the discrete shearlet transform.  This is implemented by the fast finite shearlet transform (FFST) \cite{Hauser2012}.  One level of shearlet decomposition is used, as sufficient directionality was observed at this scale.  A MATLAB implementation of the SME algorithm, part of which is also the basis of our algorithm, was described in \cite{Mallat2010} and is open source\footnote{\url{http://www.cmap.polytechnique.fr/~yu/research/SME/SME.htm}}.

\item Define a set of blocks $\mathcal{B}$.  This set is generated from a set of basic blocks, which are then rotated and translated throughout the image.  The basic blocks consist of 28 parallelograms of various areas between 12 and 18.  The set of slopes governing the orientation of these blocks are $\pm\{\frac{1}{6},\frac{1}{4},\frac{1}{3},\frac{1}{2},1,2,3,4,6\}\cup\{0,\infty\}$.  The algorithm is adaptive and more slopes can be included, to cover significant directional phenomena that are missed by the current set-up.  The blocks are translated by single pixels in the $x$ and $y$ directions throughout the image.  The rotations of the blocks allow for many edges to be isolated, exploiting the theory discussed in Section \ref{sec:math}. 

\item Compute mixing weights $\{\tilde{a}(B)\}_{B\in\mathcal{B}}$ according to the representation of $I$ in the shearlet domain by a mixed $\ell_{2}/\ell_{1}$ minimization:

\begin{align*} \tilde{a}=&\arg\min_{a}\frac{1}{2}\left\|c\left(1-\sum_{B\in\mathcal{B}}a(B)\mathbbm{1}_{B}\right)\right\|_{2}^{2}\\+&\lambda\sum_{B\in\mathcal{B}}|a(B)|\|\bar{R}_{B}c\|_{B}^{2},\end{align*}where the directional regularization factor is given by $\bar{R}_{B}c=c|_{B}-\bar{c}|_{B}$.  Here, $\bar{c}|_{B}(k,j)=$ average of the $k^{th}$ frame coefficients in $B$ located on the line passing through $j$, at angle $\theta$, where $k$ runs through all the shearlet coefficients.  In all experiments, $\lambda=.6$ \cite{Mallat2010}.  This generates a representation of the data that is simultaneously faithful and block-sparse in the coefficient space.  

\item Directional interpolators $\{U_{\theta}^{+}\}_{\theta}$ \cite{Mallat2010} and a bicubic spline interpolator $U^{+}$ are applied according to the mixing estimator as: $$\tilde{I}=U^{+}y+\sum_{\theta}(U_{\theta}^{+}-U^{+})\mathcal{SH}^{-1}\left(\sum_{B\in\mathcal{B}_{\theta}}\tilde{a}(B)c\mathbbm{1}_{B}\right).$$  

\end{enumerate}

\section{Experiments and Analysis}\label{sec:experiments}

To evaluate our SSME algorithm on noisy remote sensing images, we consider experiments on a variety of remote sensing images.  Some images had noise synthetically added, to increase the challenge of superresolution.  Noise was synthetically generated by adding independent $\mathcal{N}(0,.005)$ noise to each pixel.  This visually simulates real remote sensing noise due to properties of the sensor or poor sensing conditions.  For each experiment, the image $I$ was degraded, either by downsampling and adding noise or just downsampling, to acquire an image $\tilde{I}$.  Each superresolution algorithm under investigation was then applied to $\tilde{I}$, to acquire an approximation $\hat{I}$ to the original image $I$.  To evaluate the quality of this approximation $\hat{I}$ to $I$, we compute the PSNR and SSIM of $\hat{I}$.

\subsection{Algorithms Evaluated}

We evaluate several single-image superresolution algorithms for our experiments.  As a benchmark, we consider superresolution via bicubic interpolation \cite{Keys1981}, which is a convolutional kernel method.  This method is implemented via the MATLAB function \emph{imresize} with the `bicubic' kernel.  In addition, we compare against state-of-the-art single-image superresolution algorithms.  We consider an algorithm based on statistical models of sparse representations (SMSR) \cite{Peleg2014} and an algorithm implementing a fast alternating direction method of multipliers (ADMM) \cite{Zhao2016}.  Neither of these models require training data.  Methods based on training dictionaries and neural networks require huge amounts of training data and vast computational resources, and are in this sense incomparable to the methods considered in this article.  We also consider a directional linear estimator \cite{Mallat2010} and SME with wavelets (WSME).  

\subsection{Experimental Images}
\begin{figure}
\centering
\begin{subfigure}[t]{.32\textwidth}
\includegraphics[width=\textwidth]{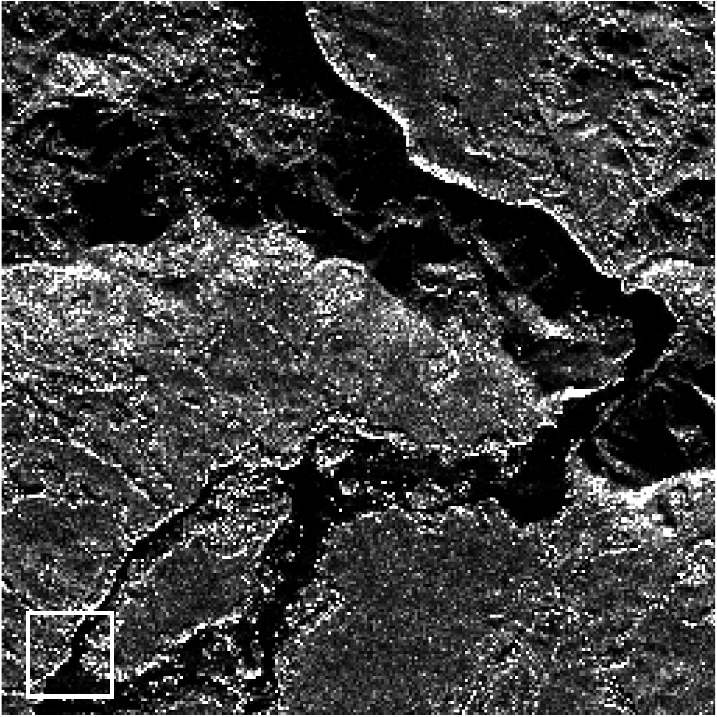}
\end{subfigure}
\begin{subfigure}[t]{.15\textwidth}
\includegraphics[width=\textwidth]{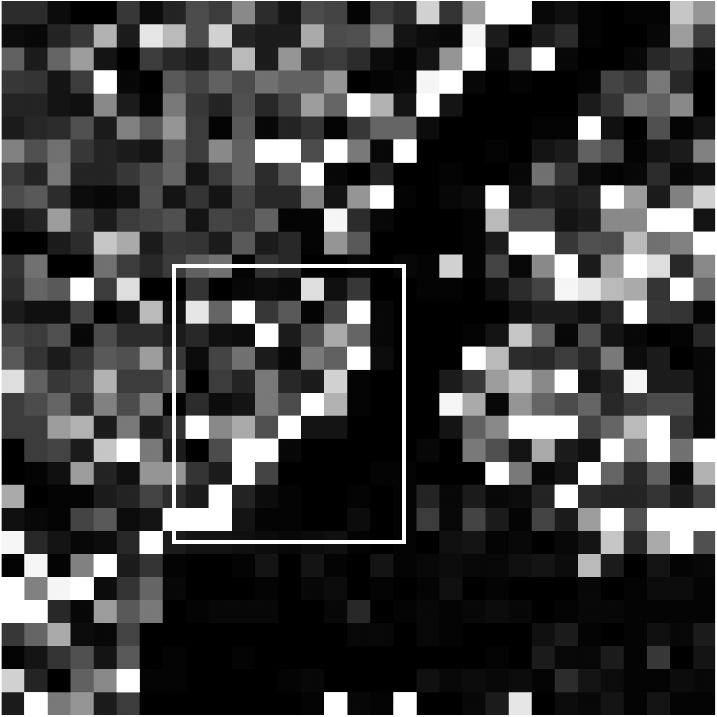}
\end{subfigure}
\begin{subfigure}[t]{.157\textwidth}
\includegraphics[width=\textwidth]{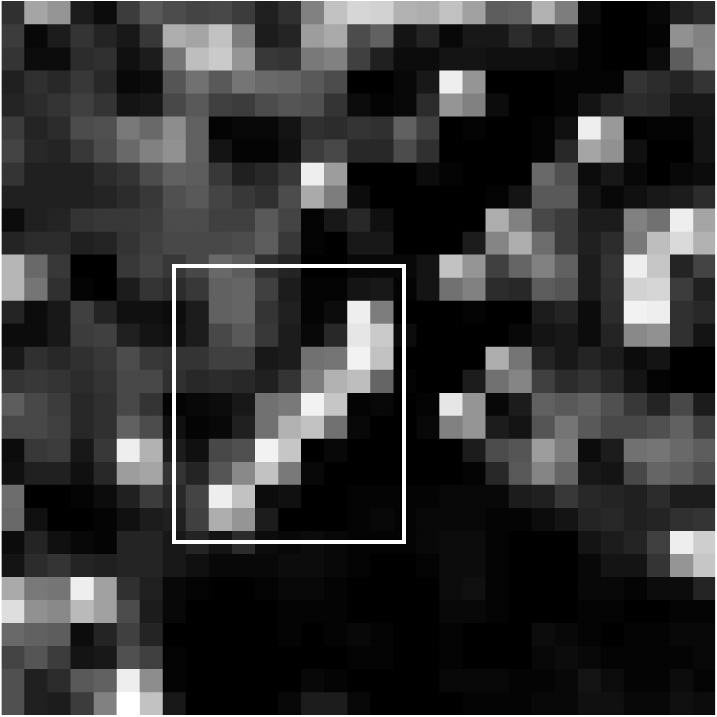}
\end{subfigure}
\begin{subfigure}[t]{.157\textwidth}
\includegraphics[width=\textwidth]{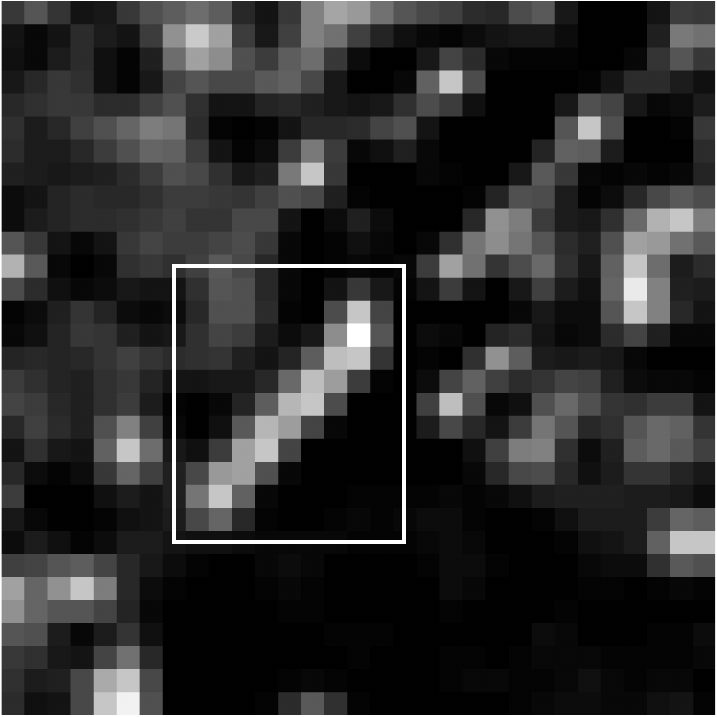}
\end{subfigure}
\begin{subfigure}[t]{.157\textwidth}
\includegraphics[width=\textwidth]{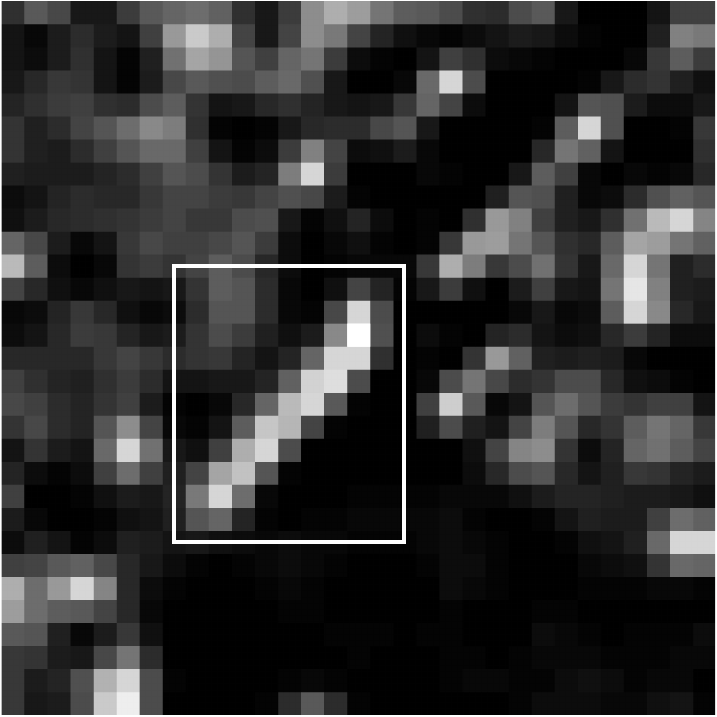}
\end{subfigure}
\caption{\label{fig:CloseUp}\emph{Top row, left to right}:  SAR image of WA, USA, with image for zoom boxed in white; zoom on original image.  \emph{Bottom row, left to right:}  zoom on ADMM output; zoom on WSME output; zoom on SSME output.  Edges are enhanced most sharply with SSME, in particular the large central edge boxed in white on all the zoomed images.}
\end{figure}

For a given test image $I$, $I$ is downsampled by a factor of 2.  Gaussian noise is then added as described above when appropriate, resulting in a corrupted image $\tilde{I}$.  We consider six images of various sizes and modalities, with some images cropped from their original size:

\begin{enumerate} 
\item A lidar derived elevation model image (DEM) ($348\times 1904$) of the University of Houston campus in TX, USA.  The image was released by IEEE Geoscience and Remote Sensing Society as part of the 2013 Data Fusion Contest.  Synthetic noise added.  
\item Band 4 ($256\times 256$) of a multispectral image of the Konza Prarie in KS, USA.  This image was captured with the ETM+ sensor, and band 4 corresponds to the near infrared electromagnetic range.  The image is courtesy of the Applied Engineering and Technology Directorate at NASA Goddard Space Flight Center.  Synthetic noise added.  
\item Band 5 ($256\times 256$) of a hyperspectral image of the Kennedy Space Center in FL, USA.  The data is open source\footnote{\url{http://www.ehu.eus/ccwintco/index.php?title=Hyperspectral_Remote_Sensing_Scenes}}.  Synthetic noise added.  
\item Band 20 ($512\times 614$) of a hyperspectral image of Cuprite in NV, USA.  The data is open source\footnotemark[\value{footnote}].  Synthetic noise added.  
\item Band 2 ($1096\times 484$) of a hyperspectral image of Pavia center in Pavia, Italy.  The data is open source\footnotemark[\value{footnote}].
\item Synthetic Aperature Radar (SAR) ($256\times256$) image of WA, USA.  The image is courtesy of the Sciences and Exploration Directorate at NASA Goddard Space Flight Center; see Figure \ref{fig:CloseUp}.
\end{enumerate}

In order to account for the random generation of noise, experiments with added noise were run 10 times, with the PSNR and SSIM averaged over these 10 trials.

\begin{figure*}
\centering
\begin{adjustbox}{max width=\textwidth}
\begin{tabular}{| c | c | c | c | c | c | c | }\hline
Image & Bicubic & SMSR & ADMM & Directional linear & WSME &  SSME\\  \hline  
Lidar &  16.9746 / 0.2615  & 16.4288 /  0.2644  & 17.1096 / 0.3161  & 17.0346 / 0.3144 & 17.1651 /  0.3209 &  \textbf{17.2503 / 0.3252 } \\ \hline 
ETM+ & 15.8717 / 0.2610  & 15.2368 / 0.2623  & 16.0060 / 0.3161 & 15.9577 / 0.3200 & 16.0840 / 0.3263 &  \textbf{16.1716 / 0.3307} \\ \hline   
KSC &   20.5967 / 0.2999 & 20.1107 / 0.2997  &  20.7357 / 0.3488 & 20.7033 / 0.3504  & 20.8326 / 0.3571  & \textbf{20.9168 / 0.3613} \\ \hline 
Cuprite &  15.8636 / 0.2767 & 15.2266 / 0.2786   & 15.9991 / 0.3309 & 15.9404 / 0.3319 & 16.0700 / 0.3383 & \textbf{16.1528 / 0.3422} \\ \hline   
Pavia HSI &  30.8707 / 0.2615 &  30.2148 / 0.2644  & 31.0684 / 0.3161 & 31.4138 / 0.3144 & 31.5683 / 0.3209 & \textbf{31.6522 / 0.3252 } \\ \hline  
SAR &  13.3796 / 0.3770  &  12.8487 / 0.3773  & 13.5442 / 0.4235 & 13.6952 / 0.4429  & 13.8467 / 0.4499  & \textbf{13.9387 / 0.4539} \\ \hline  
\end{tabular} 
\end{adjustbox}

\caption{\label{tab:Results}Numerical PSNR/SSIM results for superresolution experiments.}
\end{figure*}

As can be seen in Table \ref{tab:Results}, SSME outperforms all other methods, and in particular WSME.  This indicates the value of incorporating shearlets into superresolution of remote sensing images.  It is also of interest to observe visual improvements introduced by using shearlets instead of wavelets.  We can visually inspect the superior performance of shearlets for certain edges by zooming in on them and enhancing the contrast; see Figure \ref{fig:CloseUp}.  

\subsection{Computational Complexity}

For an $M\times N$ image, the complexity of the SME algorithm is $O(MN\log(MN))$.  The discrete shearlet transform has complexity $O(MN\log(MN))$, the same as the standard fast Fourier transform (FFT).  Thus, SSME is $O(MN\log(MN))$, making its application tractable for large images.  The computational complexity of computing shearlet features differs from computing wavelet features only by a factor logarithmic in the image size, meaning there is no substantial computational difference in using shearlets over wavelets.  
\section{Conclusions and Future Research}\label{sec:conclusion}

\subsection{Experimental Conclusions}

This article introduces the SSME superresolution algorithm, which is based on sparse mixing estimators in a shearlet frame.  Compared to other methods considered, SSME performed best in terms of PSNR and SSIM.  This is consistent with the theory of shearlets, which indicates that shearlets perform well for models of large classes of images.   The comparison state-of-the-art methods generally outperformed the benchmark methods, though WSME was the next best after SSME.  

Beyond the methods appearing in Table \ref{tab:Results}, experiments with the popular \emph{dictionary learning superresolution (DLSR)} algorithm were considered.  Machine learning 
methods are among the most sought after recent developments in superresolution; DLSR is a supervised dictionary learning algorithm that trains on images to learn filters for superresolution on new test images \cite{Yang2010}.  This method was trained using a large database of images, and did not yield strong results for the remote sensing test images considered in this article, achieving for example PSNR/SSIM values of only 13.9014/0.2386 on the ETM+ dataset. This is hypothesized to be due to the substantial differences between the training data, which consisted of a large database of optical images, and the multimodal remotely sensed testing data. This illustrates a pitfall of supervised machine learning methods: the training set must be diverse enough to capture all the variety of the test data.  Large, open source training sets for a variety of remote sensing modalities would address this problem in the future.  

\subsection{Future Research}

Discrete directional Gabor frames (DDGF) \cite{Czaja2016} are known to compress textures efficently, so it is of interest to apply them for superresolution of textures.  Additionally, joint frames of wavelets and shearlets have been proposed to separate textures from edges \cite{Elad2005,Kutyniok2009}.  Such an approach could be useful for determining which features in an image should be treated directionally with shearlets, and which isotropically with wavelets or DDGF.

\section{Acknowledgements}

The authors thank Dr. David J. Harding of the NASA Goddard Space Flight Center for supplying the SAR image; Dr. Jacqueline Le Moigne for the ETM+ image; the Hyperspectral Image Analysis group and the NSF Funded Center for Airborne Laser Mapping (NCALM) at the University of Houston for the lidar scene; and the IEEE GRSS Data Fusion Technical Committee for organizing the 2013 Data Fusion Contest.  W. Czaja was partially supported by DTRA grant 1-13-1-0015, Army Research Office grant W911NF1610008, and NSF Grant DMS 1738003.

\bibliographystyle{unsrt}
\bibliography{GRSL.bib}

\begin{thebibliography}{10}

\bibitem{Park2003}
S.C Park, M.K. Park, and M.G. Kang.
\newblock Super-resolution image reconstruction: a technical overview.
\newblock {\em {IEEE} Signal Processing Magazine}, 20(3):21--36, 2003.

\bibitem{Gu2008}
Y.~Gu, Y.~Zhang, and J.~Zhang.
\newblock Integration of spatial-spectral information for resolution
  enhancement in hyperspectral images.
\newblock {\em IEEE Transactions on Geoscience and Remote Sensing},
  46(5):1347--1358, 2008.

\bibitem{VillaChanussot2010}
A.~Villa, J.~Chanussot, J.A. Benediktsson, and C.~Jutten.
\newblock Supervised super-resolution to improve the resolution of
  hyperspectral images classification maps.
\newblock In {\em SPIE Remote Sensing Europe}, 2010.

\bibitem{Zhang2012}
H.~Zhang, L.~Zhang, and H.~Shen.
\newblock A super-resolution reconstruction algorithm for hyperspectral images.
\newblock {\em Signal Processing}, 92(9):2082--2096, 2012.

\bibitem{Liu2014}
H.~Liu, B.~Jiu, H.~Liu, and Z.~Bao.
\newblock Superresolution {ISAR} imaging based on sparse {B}ayesian learning.
\newblock {\em IEEE Transactions on Geoscience and Remote Sensing},
  52(8):5005--13, 2014.

\bibitem{Li2016}
X.~Li, F.~Ling, G.M. Foody, and Y.~Du.
\newblock A superresolution land-cover change detection method using remotely
  sensed images with different spatial resolutions.
\newblock {\em IEEE Transactions on Geoscience and Remote Sensing},
  54(7):3822--3841, 2016.

\bibitem{Wang2004}
Z.~Wang, A.C. Bovik, H.R. Sheikh, and E.P. Simoncelli.
\newblock Image quality assessment: from error visibility to structural
  similarity.
\newblock {\em IEEE Transactions on Image Processing}, 13(4):600--612, 2004.

\bibitem{Mallat2010}
S.~Mallat and G.~Yu.
\newblock Super-resolution with sparse mixing estimators.
\newblock {\em IEEE Transactions on Image Processing}, 19(11):2889--2900, 2010.

\bibitem{Guo2007}
K.~Guo and D.~Labate.
\newblock Optimally sparse multidimensional representation using shearlets.
\newblock {\em SIAM Journal on Mathematical Analysis}, 39(1):298--318, 2007.

\bibitem{compact_shearlets}
G.~Kutyniok and W.-Q. Lim.
\newblock Compactly supported shearlets are optimally sparse.
\newblock {\em Journal of Approximation Theory}, 163(11):1564--1589, 2011.

\bibitem{SR_overview}
S.C. Park, M.K. Park, and M.G. Kang.
\newblock Super-resolution image reconstruction: a technical overview.
\newblock {\em {IEEE} Signal Processing Magazine}, 20(3):21--36, 2003.

\bibitem{Czaja2014}
W.~Czaja, T.~Doster, and J.M. Murphy.
\newblock Wavelet packet mixing for image fusion and pan-sharpening.
\newblock In {\em Algorithms and Technologies for Multispectral, Hyperspectral,
  and Ultraspectral Imagery XX}, volume 9088. International Society for Optics
  and Photonics, 2014.

\bibitem{Tatem2001}
A.J. Tatem, H.G. Lewis, P.M. Atkinson, and M.S. Nixon.
\newblock Super-resolution target identification from remotely sensed images
  using a {H}opfield neural network.
\newblock {\em IEEE Transactions on Geoscience and Remote Sensing},
  39(4):781--796, 2001.

\bibitem{Peleg2014}
T.~Peleg and M.~Elad.
\newblock A statistical prediction model based on sparse representations for
  single image super-resolution.
\newblock {\em IEEE Transactions on Image Processing}, 23(6):2569--2582, 2014.

\bibitem{Foody2005}
G.M. Foody, A.M. Muslim, and P.M. Atkinson.
\newblock Super-resolution mapping of the waterline from remotely sensed data.
\newblock {\em International Journal of Remote Sensing}, 26(24):5381--5392,
  2005.

\bibitem{Murphy2016}
J.M. Murphy, J.~Le Moigne, and D.J. Harding.
\newblock Automatic image registration of remotely sensed data with global
  shearlet features.
\newblock {\em {IEEE} Transactions on Geoscience and Remote Sensing},
  54(3):1685--1704, 2016.

\bibitem{Zavorin2005}
I.~Zavorin and J.~Le Moigne.
\newblock Use of multiresolution wavelet feature pyramids for automatic
  registration of multisensor imagery.
\newblock {\em IEEE Transactions on Image Processing}, 14(6):770--782, 2005.

\bibitem{Eldar_block}
Y.C. Eldar, P.~Kuppinger, and H.~B\"{o}lcskei.
\newblock Block-sparse signals: Uncertainty relations and efficient recovery.
\newblock {\em IEEE Transactions on Signal Processing}, 58:3042--3054, 2010.

\bibitem{Christensen2003}
O.~Christensen.
\newblock {\em An Introduction to Frames and {R}iesz Bases}.
\newblock Boston: Birkh\"{a}user, 2003.

\bibitem{easley+labate+lim}
G.R. Easley, D.~Labate, and W.-Q. Lim.
\newblock Sparse directional image representations using the discrete shearlet
  transform.
\newblock {\em Applied and Computational Harmonic Analysis}, 25(1):25--46,
  2008.

\bibitem{MurphyPhD2015}
J.M. Murphy.
\newblock {\em Anisotropic Harmonic Analysis and Integration of Remotely Sensed
  Data}.
\newblock PhD thesis, University of Maryland, College Park, 2015.

\bibitem{Daubechies1992}
I.~Daubechies.
\newblock {\em Ten Lectures on Wavelets}.
\newblock Society for Industrial and Applied Mathematics, 1992.

\bibitem{Mallat1999}
S.~Mallat.
\newblock {\em A Wavelet Tour of Signal Processing}.
\newblock Academic Press, 1999.

\bibitem{Pajares2004}
G.~Pajares and J.M. De~La Cruz.
\newblock A wavelet-based image fusion tutorial.
\newblock {\em Pattern Recognition}, 37(9):1855--1872, 2004.

\bibitem{CzajaKing2012}
W.~Czaja and E.J. King.
\newblock Isotropic shearlet analogs for ${L}^{2}(\mathbb{R}^{k})$ and
  localization operators.
\newblock {\em Numerical Functional Analysis and Optimization},
  33(7-9):872--905, 2012.

\bibitem{CzajaKing2014}
W.~Czaja and E.J. King.
\newblock Anisotropic shearlet transforms for ${L}^{2}(\mathbb{R}^{k})$.
\newblock {\em Mathematische Nachrichten}, 287(8-9):903--916, 2014.

\bibitem{Negi2012}
P.S. Negi and D.Labate.
\newblock 3-{D} discrete shearlet transform and video processing.
\newblock {\em IEEE Transactions on Image Processing}, 21(6):2944--2954, 2012.

\bibitem{Easley2009}
G.R. Easley, D.~Labate, and F.~Colonna.
\newblock Shearlet-based total variation diffusion for denoising.
\newblock {\em IEEE Transactions on Image Processing}, 18(2):260--268, 2009.

\bibitem{Murphy2015}
J.M. Murphy and J.~Le Moigne.
\newblock Shearlet features for registration of remotely sensed multitemporal
  image.
\newblock In {\em Proceedings of IEEE International Conference Geoscience and
  Remote Sensing Symposium}, 2015.

\bibitem{Murphy2016_Agile}
J.M. Murphy, O.N. Leija, and J.~Le Moigne.
\newblock Agile multi-scale decompositions for automatic image registration.
\newblock In {\em Algorithms and Technologies for Multispectral, Hyperspectral,
  and Ultraspectral Imagery XXII}, volume 9840, page 984011. International
  Society for Optics and Photonics, 2016.

\bibitem{Miao2011}
Q.~Miao, C.~Shi, P.-F. Xu, M.~Yang, and Y.-B. Shi.
\newblock A novel algorithm of image fusion using shearlets.
\newblock {\em Optics Communications}, 284(6):1540--1547, 2011.

\bibitem{CzajaMurphy2015}
W.~Czaja, J.M. Murphy, and D.~Weinberg.
\newblock Superresolution of remotely sensed images with anisotropic features.
\newblock In {\em Proceedings of SAMPTA}, 2015.

\bibitem{Bosch2015}
E.H. Bosch, W.~Czaja, J.M. Murphy, and D.~Weinberg.
\newblock Anisotropic representations for superresolution of hyperspectral
  data.
\newblock In {\em SPIE Defense+ Security. International Society for Optics and
  Photonics}, 2015.

\bibitem{Hauser2012}
S.~H\"{a}user.
\newblock Fast finite shearlet transform.
\newblock arXiv:1202.1773, 2012.

\bibitem{Keys1981}
R.~Keys.
\newblock Cubic convolution interpolation for digital image processing.
\newblock {\em IEEE Transactions on Acoustics, Speech and Signal Processing},
  29(6):1153--1160, 1981.

\bibitem{Zhao2016}
N.~Zhao, Q.~Wei, A.~Basarab, N.~Dobigeon, D.~Kouam{\'e}, and J.~Tourneret.
\newblock Fast single image super-resolution using a new analytical solution
  for $\ell_2-\ell_2$ problems.
\newblock {\em IEEE Transactions on Image Processing}, 25(8):3683--3697, 2016.

\bibitem{Yang2010}
J.~Yang, J.~Wright, T.~Huang, and Y.~Ma.
\newblock Image super-resolution via sparse representation.
\newblock {\em IEEE Transactions on Image Processing}, 19(11):2861--2873, 2010.

\bibitem{Czaja2016}
W.~Czaja, B.~Manning, J.M. Murphy, and K.~Stubbs.
\newblock Discrete directional {G}abor frames.
\newblock {\em To appear in Applied and Computational Harmonic Analysis}, 2016.

\bibitem{Elad2005}
M.~Elad, J.-L. Starck, P.~Querre, and D.~L. Donoho.
\newblock Simultaneous cartoon and texture image inpainting using morphological
  component analysis ({MCA}).
\newblock {\em Applied and Computational Harmonic Analysis}, 19(3):340--358,
  2005.

\bibitem{Kutyniok2009}
G.~Kutyniok and D.~Labate.
\newblock Resolution of the wavefront set using continuous shearlets.
\newblock {\em Transactions of the American Mathematical Society},
  361(5):2719--2754, 2009.

\end{thebibliography}

\end{document}